\pdfoutput=1

\documentclass[11pt]{article}
\usepackage{graphicx}
\usepackage{ACL2023}
\usepackage[multiple]{footmisc}
\usepackage{longtable}
\usepackage{times}
\usepackage{latexsym}
\usepackage{hyperref}
\usepackage{amsmath}
\usepackage[T1]{fontenc}

\usepackage[utf8]{inputenc}

\usepackage{microtype}

\usepackage{inconsolata}

\title{Mental Disorder Classification via Temporal Representation of Text}

\author{Raja Kumar$^\ast$,  Kishan Maharaj$^\ast$,  Ashita Saxena,  Pushpak Bhattacharyya \\
Indian Institute of Technology Bombay, Mumbai, India \\
   \texttt{\{kumar.raja.iitb, kishan.maharaj.iitb, as.saxena.as\}@gmail.com,} \\
      \texttt{
    pb@cse.iitb.ac.in}}

\begin{document}
\maketitle
\def\thefootnote{$\ast$}\footnotetext{Equal Contributions}\def\thefootnote{\arabic{footnote}}
\begin{abstract}



Mental disorders pose a global challenge, aggravated by the shortage of qualified mental health professionals. Mental disorder prediction from social media posts by current LLMs is challenging due to the complexities of sequential text data and the limited context length of language models. Current language model-based approaches split a single data instance into multiple chunks to compensate for limited context size. The predictive model is then applied to each chunk individually, and the most voted output is selected as the final prediction. This results in the loss of inter-post dependencies and important time variant information, leading to poor performance. We propose a novel framework which first compresses the large sequence of chronologically ordered social media posts into a series of numbers. We then use this time variant representation for mental disorder classification. We demonstrate the generalization capabilities of our framework by outperforming current SOTA in three different mental conditions: \textit{depression}, \textit{self-harm}, and \textit{anorexia} by an absolute improvement of 5\% in F1 score. We also investigate the situation when current data instances fall within the context length of language models and present empirical results highlighting the importance of temporal properties of textual data. Furthermore, we utilize the proposed framework for a transfer-learning study, exploring commonalities across disorders and the possibility of inter-domain data usage.


\end{abstract}

\section{Introduction}
\label{sec:introduction}

In the contemporary world, mental health plays a crucial role in a person's overall well-being. The World Health Organization (WHO)\footnote{\url{https://www.who.int/news-room/fact-sheets/detail/mental-disorders}}  highlights the intensity of this matter by reporting that globally, one in every eight individuals suffers from a mental disorder. A comprehensive study \cite{MCGRATH2023668} reveals that over 50\% of people worldwide confront a mental health issue at some point in their lives. The scarcity of adequately trained professionals hinders access to timely and effective intervention, motivating the need for automated mental disorder detection. 

\begin{figure}[hbt!]
     \centering
    \includegraphics[width=0.46\textwidth]{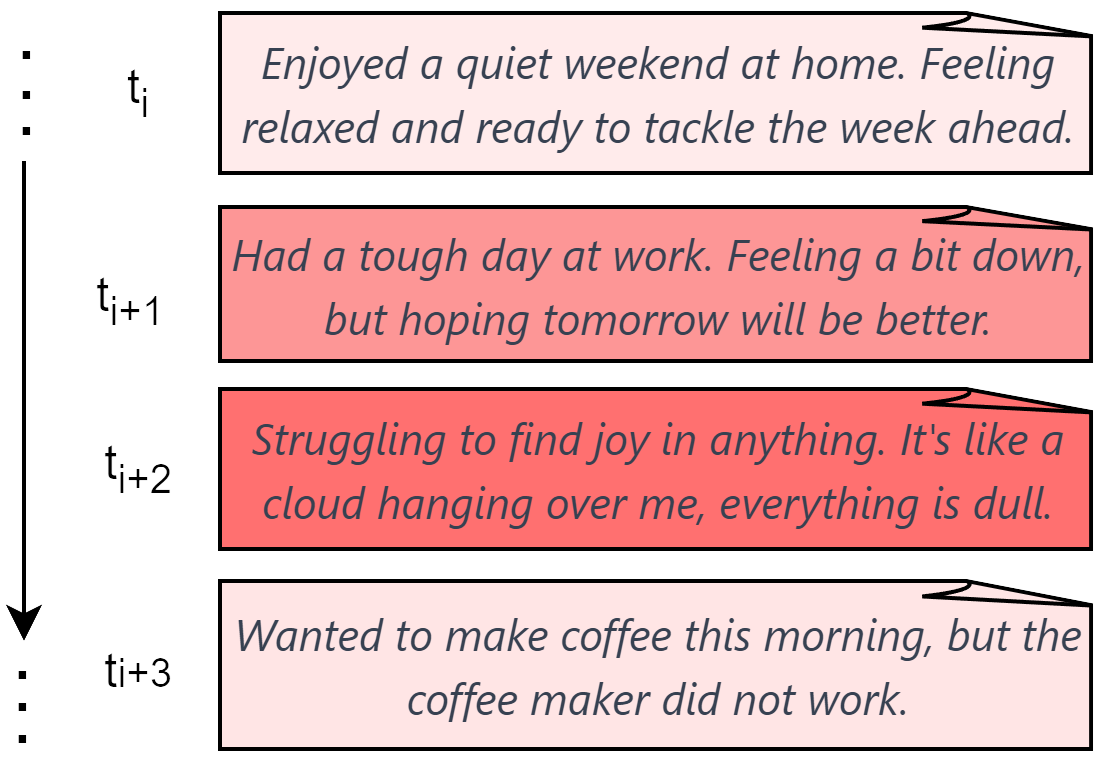}
    
   \caption{An example of four posts made by a person on social media. The intensity of the red colour indicates the extent to which a post indicates depression.}
	\label{fig:temporal_posts}
\end{figure}

\begin{figure*}
    \centering
    \includegraphics[width=1\textwidth]{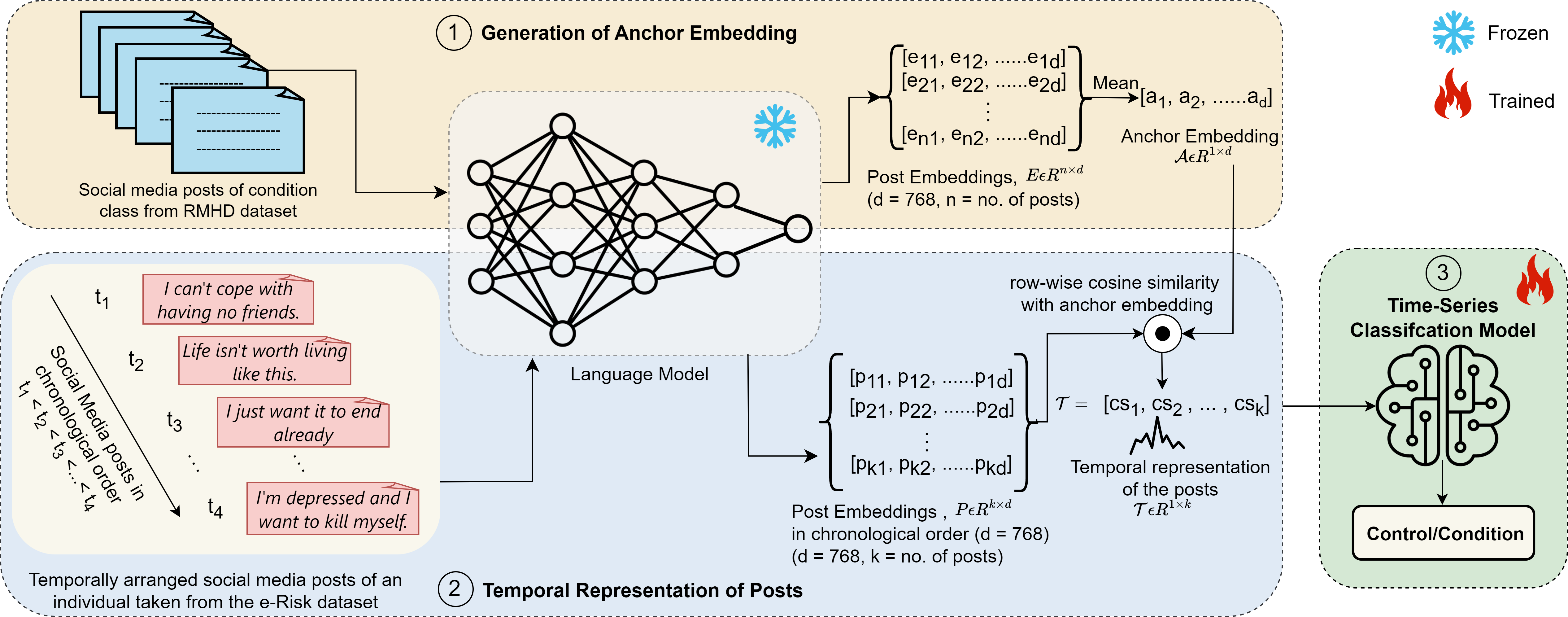}
    \caption{This figure shows the overall pipeline of our approach. Here, (1) shows the generation of the anchor embedding from RMHD, (2) shows the creation of temporal representations of social media posts of an individual, and (3) depicts the classification of the temporal representations as control or condition. Generating the anchor embedding is the first step which is followed by representing the posts in a temporal manner. These temporal representations are then used to train the time series classification model to detect the presence of a disorder.}
    \label{fig:time-series-representation-method}
\end{figure*}

Social media platforms, like Reddit, have become a widespread outlet for self-expression, thus becoming a source of user-generated content that may provide valuable insights into individuals' mental states. The mental state of a patient varies from time to time, depending on the severity of symptoms.
Figure \ref{fig:temporal_posts} shows an example of a situation with such behavioural variation. The post at time $t_{i+2}$ shows a peak in depression in the person, whereas the post at time $t_{i+1}$ can be seen as the onset of depression. Posts at time $t_i$ and $t_{i+3}$ are indicative of the general behaviour of the person. These rise and falls in the sentiment polarity of social media posts represent temporal properties that can serve as important information for detecting mental disorders. These clues can be used to differentiate patients with mental disorders from healthy subjects efficiently.  



The central idea of recent language model-based approaches (\citealp{aragon-etal-2023-disorbert, ji-etal-2022-mentalbert}) is to concatenate multiple texts to form a chunk and then apply a predictive model to each chunk separately. The final prediction is positive if most chunks are classified as positive. 

The major issues in such approaches are: 1) \textbf{Loss of temporal information}: applying the predictive model on chunks separately leads to the loss of order and inter-post dependencies. 2) \textbf{Lack of global view}: the concentration of positive class text in only a few of the chunks can lead to misclassification because of majority voting. 3) \textbf{Semantic noise}: merging posts to form a single chunk with different semantics can diminish the significance of a single post and introduce noise in the data. For instance, a post indicating severe depression could be concatenated with the subsequent post discussing the positive outcomes after the treatment.



To handle these issues, we must account for two aspects: 1) preserving the post-identity and order and 2) global classification i.e. classifying the entire sequence of posts together instead of classifying chunks of social media text. Using current LLMs here can be challenging due to computational costs and the high context length of social media posts. For instance, the average context length of a subject is 11.6K words (max 89.4K) Table \ref{tab:data_statistics}, which exceeds the input context length of current LLMs. 

Hence, the chunking-based methods can not address these concerns, while other LLM-based approaches may prove infeasible within the given context. This motivates the need for a representation technique capable of compressing information while retaining the time-variant properties of social media posts.
We build upon these insights and propose a framework to account for these issues. We perform rigorous analyses to establish the importance of temporal properties in mental disorders (Section \ref{sec:temporal_analysis}). We also discuss the performance of the current language models when the input length is within the capacity of the language model (Section \ref{sec:full-context}). Our source code is available on GitHub\footnote{\href{https://github.com/raja-7-c/Mental-Disorder-Classification-via-Temporal-Representation-of-Text.git}{GitHub Link}} for academic purposes.

Our contributions are:
\begin{itemize}
    \item A representation method for mental-health domain which \textit{compresses textual data from social media posts into a time series format} to capture the time-dependent patterns of a patient (Figure \ref{fig:time-series-representation-method}). This provides a temporal representation of the textual data while reducing the floating point operations (FLOPs) by at least 330 times (Table \ref{tab:flops_comparison}) compared to SOTA.

    \item A novel framework incorporating temporal data for mental disorder identification by using foundational deep learning models. Our approach outperforms language model-based methods by 5\% in F1 score (Table \ref{tab:ERISK-RESULTS}) across three mental conditions: anorexia, depression, and self-harm.

    \item A transfer-learning study (Section \ref{sec:cross-domain}) of the three disorders to understand the commonality across disorders. We investigate the possibility of cross-domain data usage (Table \ref{tab:cross_domain_results}), which can further benefit the identification of low-resource mental disorders.  
\end{itemize}

\noindent
In the subsequent sections, we discuss recent works that are relevant to our contributions. We then discuss the dataset used and the proposed method, followed by the insights from our experiments.

\section{Related Work}

In this work, \textbf{\textit{control}} class refers to the class of healthy individuals not diagnosed with a psychiatric disorder whereas \textbf{\textit{condition}} class refers to the class of individuals diagnosed with a psychiatric disorder. For our study, we focus on two mental disorders, anorexia and depression, as well as a mental condition, self-harm. For brevity, we collectively refer to them as mental disorders (like \citealp{aragon-etal-2023-disorbert, ji-etal-2022-mentalbert}).
    

Mental disorder prediction from social media has seen significant development in the last decade. Early works transitioned from the use of low-level handcrafted features like Linguistic Inquiry and Word Count (LIWC) (\citet{islam2018depression}; \citet{shrestha2019detecting}; \citet{simms2017detecting}) to high semantic features like word or document embeddings \citep{friedenberg2016umd, bandyopadhyay2019identification, lin2017disorder, hemmatirad2020detection}. This was succeeded by representation learning-based approaches \citep{rao2020mgl, wongkoblap2019predicting, gaur2021characterization}, which operate on user-level prediction and eliminate the need for explicit feature engineering. 

Other works leverage longitudinal data to capture unique patterns of emotional transitions shown by mental patients. These approaches use chunking to process $m$ words (\citet{trotzek2018utilizing}; \citet{uban2021emotion}; \citet{orabi2018deep}) or $n$ posts (\citet{ragheb2019attentive}; \citet{mitchell2015quantifying}) sequentially and perform classification using majority voting. An alternative method involves feature extraction by concatenating all posts (\citet{aguilera2021depression}; \citet{jamil2017monitoring}) related to a specific subject. However, these approaches fail to incorporate the temporal variations between the posts of a subject because of the usage of chunking and majority voting (see Section \ref{sec:introduction}). 




A few studies closely align with our approach to constructing temporal representations of social media posts. \citet{reece2017forecasting} employed state-space temporal analysis using day and week as the time window for depression detection. \citet{de2013predicting} examined a user's tweets within a single day to derive various behavioural measures and constructed a time series for each measure. \citet{chen2020examining} created a time series representation of the mood profile using traditional sentiment retrieval models. \citet{lee2023towards} proposed a multi-task framework aimed at predicting suicidality for bipolar disorder patients by categorizing individual posts into different suicidality levels. \citet{sawhney2021suicide} introduced graph-based approaches to model user interaction and the temporality of posts, while \citet{sawhney2021phase} modeled "phases" at the post level, requiring post-level annotations for suicide ideation detection. A significant limitation of these approaches is either their reliance on low-level features, which fail to capture the deeper semantic understanding of emotional aspects in human language, or the need for user interaction data and post-level annotations, which are not available in our setup.

\citet{guo2021emotion} fine-tuned BERT to generate emotional post representations and derived emotion transition matrices. However, emotional states were extracted individually from each post, overlooking inter-post dependencies. Another drawback is their reliance on the first-order Markov assumption for prediction.
Recent works like \citet{ji-etal-2022-mentalbert} pre-trained a BERT model on the social media and 
 mental health dataset. \citet{aragon-etal-2023-disorbert} performed a double domain adaptation (extending the pre-training process) on BERT, incorporating data from Reddit and mental health sources. These approaches again rely on chunking and majority voting while facing the limitation of high expense.

In this work, we propose a novel framework which captures the temporal patterns of social media posts without the use of majority voting and chunking.

\begin{table*}[htb!]
    \centering
    \small
    \begin{tabular}{cccccccc}
    \hline
     &  \multicolumn{2}{c}{Training}& \multicolumn{2}{c}{Validation}& \multicolumn{2}{c}{Test}& \multicolumn{1}{c}{Total}\\
    \hline
         &  Condition&  Control&  Condition&  Control&  Condition&  Control& \\
    \hline
    &&&\textbf{Anorexia}\\
    \hline
    \#subjects &  45 &  332 & 14 & 81 &  73& 742 & 1307\\
    avg \# posts&  404.7 &  552.3 & 411.9 &  560.9 &  241.4 & 745.1 & 639.44\\
    avg \# words&  36.2&  21.1 & 39.6&  20.9&  37.2& 21.7 & 23.10\\
    \hline
    &&&\textbf{Depression}\\
    \hline
     \#subjects&  173&  1195 & 44 &  298 & 40& 49 & 1799\\
    avg \# posts&  444.9&  663.4 &  436.7&  658.2&  493.0& 543.7 & 629.31\\
    avg \# words&  24.2&  20.55 &  29.8&  24.77 &  39.2& 45.6 & 22.91\\
    \hline
    &&&\textbf{Self-Harm}\\
    \hline
    \#subjects&  29 &  243 & 12 &  56 & 104& 319 & 763\\
    avg \# posts&  172.0&  543.9 &  167.8&  549.7&  112.4& 285.6 & 357.52\\
    avg \# words&  22.4&  17.5 &  26.8&  19.7 &  21.4& 11.9 & 16.17\\
    \hline
    \end{tabular}
								
    \caption{Statistics of the e-Risk datasets for anorexia, depression and self-harm. The \textit{control} class refers to the class of healthy individuals, and the \textit{condition} class refers to the class of individuals diagnosed with a disorder.}
    \label{tab:data_statistics}
\end{table*}

								

\section{Datasets}

We use two datasets in this work: eRisk evaluations and Reddit Mental Health Dataset (RMHD).

\paragraph{e-Risk evaluation datasets:} We utilize datasets from the e-Risk\cite{losada2016test} evaluation for anorexia \cite{losada2019overview}, depression \cite{losada2018overview}, and self-harm \cite{losada2020overview} in the given splits. These datasets consist of the post history of Reddit users. Depression labels were obtained by thresholding \cite{beck1961inventory} on the Beck Depression Inventory scores. We obtain validation data by randomly sampling 20\% of the train set (refer to Table \ref{tab:data_statistics} for statistics).



\paragraph{Reddit Mental Health Dataset (RMHD):} We use the Reddit Mental Health Dataset (RMDH) \cite{low2020natural} to generate the anchor embeddings (Section \ref{sec:reference_embedding}) of anorexia, depression and self-harm (refer Table \ref{tab:RMHD} for statistics).


\section{Methodology}
\label{sec:Time Series Representation}
\paragraph{Problem Formulation:}
Given social media posts $ \{p_1, p_2, \ldots, p_k\}$ by a subject $\mathcal{P} $ in a chronological order, we first obtain the time-series representation $\mathcal{T} = \{t_1, t_2, \ldots, t_k\}$ of this textual data. We then detect the presence of a mental disorder by treating this as a binary time-series classification problem.

A naive approach for the above could be a simple concatenation of these posts sequentially followed by the classification via an LLM in a single pass. However, this is not feasible in our use case as there are a large number of words per subject. For instance, we have an average of 14.4K words per subject (max 89.4K) for depression. This exceeds the maximum token length handled by contemporary language models (refer to Appendix \ref{sec:appendix_context_lengths}). Consequently, there is a requirement for a representation technique that can condense this extensive information into more compact data units for effective processing while preserving the temporal aspects of the data. 
Intuitively, the process of mental disorder detection involves determining whether a subject's text resembles that of someone who is diagnosed with the disorder. This involves comparing the social media posts made by a particular subject with those of the condition class. We aim to imitate this logic in our proposed technique. 
We further elaborate our methodology in the following subsections.

\subsection{Anchor Embedding Generation}
\label{sec:reference_embedding}
The initial step involves creating an anchor embedding which serves as a semantic anchor of a specific disorder's condition class. A similar approach was used by \citet{fei2015social} to address covariate shifts in social media text classification.

For this, we first fetch the sentence embeddings for every post of a particular disorder from the RMHD Dataset \cite{low2020natural}. This is done by using the frozen MPNet model \cite{song2020mpnet}, which is trained on contrastive loss aiming to learn semantic similarity \cite{reimers2019sentence}.
The embedding representations of all the posts belonging to a disorder are then aggregated by using the mean operation (part 1 of Figure \ref{fig:time-series-representation-method}). We call this aggregated representation as the \textit{anchor embedding}, $\mathcal{A}$.  
For a disorder $\mathcal{D}$, if $E_1, E_2, \ldots, E_n$ are the $n$ embedding vectors for $n$ social media posts of condition class of a disorder, then the anchor embedding is calculated as:
\begin{center}
\vspace{-1.5pt} 
$\mathcal{A_{D}} = \frac{1}{n} \sum_{i=1}^{n} E_i$    
\vspace{-1.5pt} 
\end{center}

These anchor embeddings are then used to obtain the temporal representation of the textual data. To avoid data leaks, we obtain the anchor embedding from the Reddit Mental Health Dataset (RMHD) \cite{low2020natural} and report the results on the e-Risk benchmark. For the anchor embedding of self-harm, we used posts from the suicide subreddit of RMHD as there is a direct association between self-harm and suicide attempts \cite{duarte2020self}.

\subsection{Time Series Representation}
After generating the anchor embedding, we use it to obtain the temporal representation of the textual data. 
To do this, we take the social media posts of an individual in chronological order and calculate the cosine similarity score between the embedding of each post and the anchor embedding. This generates the temporal representation for a user as explained in Figure \ref{fig:time-series-representation-method} (part 2). 
We hypothesize that semantic similarity between the subject under consideration and the condition class is more crucial than the absolute semantic information of the subject itself for effectively classifying mental disorders.




\begin{figure}
    \centering
    \includegraphics[width=0.47\textwidth]{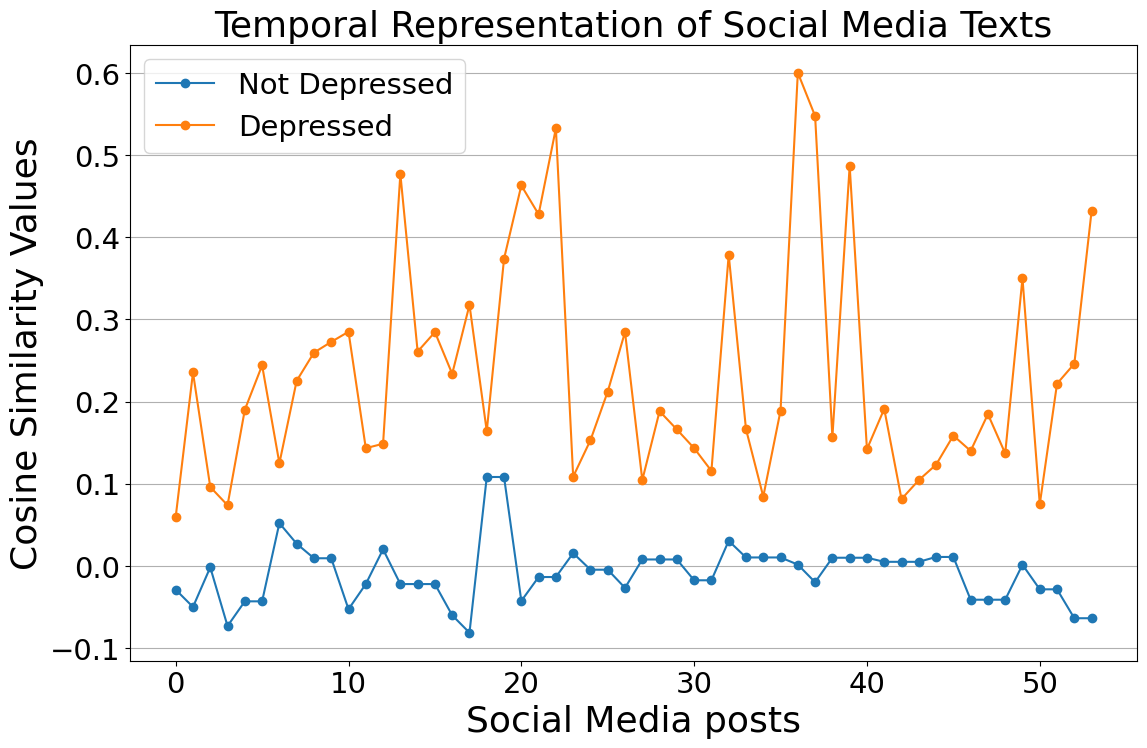}
    \caption{Temporal representation of depressed and non-depressed subject. The Y-axis is the cosine similarity value with the anchor embedding, and the X-axis is the posts arranged according to the time of posting.}
    \label{fig:temporal_representation}
\end{figure}

For a particular disorder $\mathcal{D}$, let $P_1, P_2, \ldots, P_k$ be the $k$ post embedding vectors of a particular subject in e-Risk data. Here, $P_j$ denotes the sentence embedding vector of a post at the $j$-th time step $(1 \leq j \leq k)$ for that individual subject. We calculate the cosine similarity $cs_j$ between the reference embedding $\mathcal{R_{D}}$ of that disorder and each post embedding $P_j$.

All $cs_j$ vectors from $j = 1$ to $k$ represent the time series data $\mathcal{T} = \{cs_1, cs_2, \ldots, cs_k\}$ for a particular subject representing the social media post in temporal format. 
Therefore, each post is represented as a scalar value indicating its distance from the anchor embedding. 

Since the anchor embedding is calculated from the condition class, the cosine similarity is expected to show low values if the post is from the control class and high values if the post is from the condition class. As shown in Figure \ref{fig:temporal_representation}, this comparison of the temporal representation for a depressed and non-depressed subject highlights a clear distinction between the two classes. This demonstrates the capability of our representation methodology.

\begin{table*}[htbp!]
    \centering
   \small
    \begin{tabular}{ccccccccccc}
    \hline
Method &  &\multicolumn{3}{c}{Anorexia} & \multicolumn{3}{c}{Depression} & \multicolumn{3}{c}{Self-Harm} \\
    \hline
    & Masking & F1 & P & R & F1 & P & R & F1 & P & R \\
    \hline
    \multicolumn{11}{c}{Baselines} \\
    \hline

    
    BERT & Random & 0.77 & 0.70 & 0.85 & 0.62 & 0.55 & 0.72 & 0.60 & 0.44 & 0.94 \\
    MentalBERT & Random & 0.76 & 0.66 & 0.89 & 0.67 & 0.57 & 0.80 & 0.71 & 0.62 & 0.84 \\
   
    BERTw/Reddit &  Random &0.81 & 0.75 & 0.88 & 0.66 & 0.56 & 0.80 & 0.71 & 0.66 & 0.76 \\
    
    BERTw/Reddit & Guided & 0.82 & 0.82 & 0.82 & 0.68 & 0.55 & 0.90 & 0.72 & 0.65 & 0.82 \\
    
    BERTw/Health & Random & 0.80 & 0.77 & 0.84 & 0.67 & 0.53 & 0.93 & 0.69 & 0.60 & 0.82 \\
    
    BERTw/Health & Guided & 0.82 & 0.81 & 0.84 & 0.68 & 0.57 & 0.85 & 0.74 & 0.72 & 0.76 \\

    DisorBERT & Random &  0.82 & 0.83 & 0.81 & 0.68 & 0.54 & 0.93 & 0.72 & 0.65 & 0.80 \\
    
    DisorBERT &  Guided & \textbf{0.83} & 0.82 & 0.85 & 0.69 & 0.56 & 0.89 & 0.72 & 0.73 & 0.71 \\
    MPNetv2 (ZS){*} &  - & 0.16 & 0.09 & 1.00 & 0.62 & 0.45 & 1.00 & 0.40 & 0.25 & 1.00 \\

    MPNetv2 (FT [eRisk]){*} &  - & 0.71 & 0.60 & 0.89 & 0.62 & 0.57 & 0.68 & 0.48 & 0.89 & 0.33 \\

    MPNetv2 (FT [eRisk+RMHD]){*} &  - & 0.78 & 0.73 & 0.85 & 0.62 & 0.45 & 1.00 & 0.42 & 0.27 & 0.98 \\

    GPT-3.5-turbo{*} &  - & 0.05 & 1.00 & 0.03 & 0.37 & 1.00 & 0.23 & 0.22 & 0.93 & 0.12 \\

    MentalLLaMA-chat-13B{*} & - & 0.08 & 1.00 & 0.04 & 0.05  &  1.00 & 0.03 & 0.02 &  0.50 & 0.01 \\
    
    \hline
    \multicolumn{10}{c}{Our Methods} \\
    \hline

    




    

    




    
    Feedforward Network & - &  \textbf{0.83} & 0.87 & 0.79 & 0.71 & 0.83 & 0.59 & 0.81 & 0.84 & 0.78 \\

    1D-CNN & - &  0.82 & 0.86 & 0.78 & 0.70 & 0.77 & 0.65 & \textbf{0.83} & 0.85 & 0.81 \\

    




    
    LSTM & - & 0.79 & 0.84 & 0.74 & \textbf{0.75} & 0.79 & 0.71 & \textbf{0.83} & 0.93 & 0.75 \\
    
    Transformer & - &  0.82 & 0.85 & 0.79 & 0.71 & 0.83 & 0.61 & 0.74 & 0.81 & 0.67 \\

    




    
    \hline
    \end{tabular}

\caption{F1, precision (P), and recall (R) values over the condition class in three eRisk tasks: anorexia, depression and self-harm. “BERTw/Reddit” indicates the model is only fine-tuned on Reddit texts, and “BERTw/Health” is only fine-tuned on mental health datasets. ZS and FT refer to Zero-Shot and Fine-Tuned experiments respectively. Bold numbers denote the best F1 score of a particular disorder across all methods. The results of our methods have been reported after averaging over five random seeds. The first eight baseline values were taken from \citet{aragon-etal-2023-disorbert}, and the baselines marked with * were trained by us.} 
\label{tab:ERISK-RESULTS}
\end{table*}

\subsection{Time Series Classification}
\label{subsec:Time-series-Classification}

For time series classification  (Figure \ref{fig:time-series-representation-method} part (3)), we explore two approaches: (i) Feature extraction-based, where we train a feed-forward network on statistical and temporal features extracted from the time series data, using the top 30 features for classification. This selection of raw time series features was motivated by our interest in assessing the performance of a basic feature engineering approach for this task. We extracted these features to capture relevant characteristics from the data.
(ii) Representation learning-based, where we directly feed the raw time series data as input to the model.
For representation learning-based approaches, we use three different architectures, namely: 1) LSTM \cite{yu2019review} 2) 1D CNN \cite{tang2020rethinking}, and 3) Time-Series Transformer \cite{zerveas2021transformer}.
We use the method of threshold moving \cite{zhang2020threshold} to decide the best probability threshold for the decision boundary to account for data imbalance (Table \ref{tab:data_statistics}). We describe all the experimental details including features list in the Appendix \ref{sec:appendix-model-config-baseline}.

\section{Baseline Approaches}
\label{sec:baseline_approaches}
We use all the baseline approaches introduced by \citet{aragon-etal-2023-disorbert} for comparing our approach.
MentalBERT \cite{ji-etal-2022-mentalbert} pre-trained BERT on social media and mental health text, and DisorBERT \cite{aragon-etal-2023-disorbert} applied double-domain adaptation with random and guided masking.

Additionally, we use MPNet \cite{song2020mpnet} as a baseline in both zero-shot and fine-tuned settings. To compare the results with current SOTA language models, we report results on GPT3.5 turbo \cite{brown2020language} and MentalLLaMA-chat-13B \cite{yang2024mentallama} instruction tuned on large mental health data.  
We follow \citet{aragon-etal-2023-disorbert} for evaluation and training. All the baseline methods used the chunking (k=35) and majority voting-based approach. The F1 score of the positive class serves as our standard metric for assessing the experimental results. We provide a detailed explanation for baselines in Appendix \ref{sec:baseline_appendix}
and present experimental results on additional baselines in Table \ref{tab:additional_exp}, Appendix \ref{sec:appendix-additional-exp}.


%

\section{Results and Observations}
\label{sec:Results-and-Observation}
The results of our experiments and the baseline results are summarized in Table \ref{tab:ERISK-RESULTS}. Our approach shows an improvement of 5\% over the SOTA averaged over all three tasks. 
We observe an absolute improvement of 9\% and 6\% in the F1 score for self-harm and depression respectively over the current SOTA.
For anorexia, our model has the same F1 score as the SOTA. These results indicate the benefits of temporal representation of data and global classification. 
We perform an in-depth analysis in Section \ref{sec:pred_analysis} and share insights into the predictive behaviour of our approach.

We observe (Table \ref{tab:ERISK-RESULTS}) very low F1 scores with MPNet (zero-shot) because most instances are classified as positive (high false positive). We observe improvements after fine-tuning MPNet with eRisk data over the zero-shot strategy but not more than the current SOTA. We also fine-tune MPNet on both eRisk and RMHD datasets. We note that the performance in detecting self-harm and depression is significantly worse compared to the proposed method, even though we have provided RMHD data explicitly by fine-tuning. This suggests that just fine-tuning a model on mental health data is insufficient for effective mental disorder classification indicating the importance of temporal information and complete context. This further demonstrates the prowess of our representation technique which extracts the embedding without fine-tuning.

The poor performance of GPT-3.5-turbo and MentalLLaMA-chat-13B (Table \ref{tab:ERISK-RESULTS}) may be due to the chunking strategy, which does not consider the longitudinal history of the users's texts and removes the inter-post dependencies. 
Since important signals are concentrated in some of the posts, most of the posts may not have sufficient information for accurate mental disorder prediction.
We provide additional insights with an example in Appendix \ref{sec:poor-performance-gpt3.5-mentalLLaMa}. To go one step further in this analysis, we present the results on the instances when the text is within the range of maximum input size in Section \ref{sec:full-context}. 




\section{Analysis}


\label{sec:pred_analysis}
Our approach demonstrates improvement in self-harm and depression prediction while showing a competitive performance in anorexia prediction (Table \ref{tab:ERISK-RESULTS}). This highlights the necessity of considering the complete context of the input data and its temporal information. Our methodology differs from other approaches that focus on understanding the absolute semantics of a subject. Instead, we model the semantic contrast between the posts of the subject and the anchor embedding representing the condition class. This approach enhances our ability to identify disorders characterized by high semantic contrast, like self-harm.





The social media posts of individuals diagnosed with self-harm use more provocative and intense language compared to posts about other mental conditions. For example, 
\textit{"I sliced my arm with a box cutter. It almost seeped through my long sleeve since it bled so much."} These strong linguistic indications cause a huge contrast between the control subject's posts and the condition subject's posts, leading to a 9\% improvement over SOTA. 

The low improvement in anorexia can be attributed to the high noise in the test set. In most posts by the condition class, individuals have a lot of irrelevant information. For example, \textit{"To be honest, I have no problem paying a small bit extra for Amazon. Their client service is really great."} 
These kinds of instances result in noisy test data, resulting in a low improvement over SOTA. 




\subsection{Error Analysis}
\label{sec:error-analysis}
We discuss two major error scenarios in the predictions made by our best-performing models.

\paragraph{Out-of-context posts:} There are many social media posts from condition-class individuals that may not provide sufficient information for predicting the accurate mental states of the individual. Consider the posts from a depressed individual:

\noindent
1) \textit{Can the chip on XZ Genco ES be repaired?} 

\noindent
2) \textit{My cat's reaction to not living alone anymore}

\noindent
3) \textit{This BMW got 30\% cooler in a few seconds.}



The information conveyed by the above sentences is unrelated to the mental well-being of the individual and does not contain any clues to convey the depressed mental state. This results in the misclassification of this instance as non-depressed. 
This scenario could also be observed in individuals facing mental health challenges who may refrain from expressing their emotions and thoughts due to the prevalent social stigma. 

\paragraph{Incomplete context:} We observe a few cases in which only a limited number of posts (< 10) are available for a particular individual. This is seen when the new social media users have not posted enough to gather context related to their mental health. Also, for some condition subjects, this may be due to the Reddit policies on NSFW (Not Safe for Work) content. Since most of the posts in mental forums are highly triggering, they are automatically removed by the policy-enforcing bots. This resulted in incomplete context for some users.

\subsection{Efficiency Analysis}
\label{sec:efficiency-analysis}

\begin{table}
    \centering
    \small
    \begin{tabular}{p{2cm}p{1.3cm}p{1.3cm}p{1.3cm}}
    \hline
    \textbf{Model} & \textbf{Anor} & \textbf{Depr} & \textbf{SH} \\
    \hline
    Feedforward & 2.47 K & 2.51 K  & 1.89 K \\
    LSTM &  4.18 K & 4.23 K  & 4.30 K  \\
    1D-CNN &  25.44 M & 5.87 M &  5.85 M \\
    Transformer & 14.61 M & 14.62 M & 14.56 M \\
    BERT{*} & 8.42 B & 8.42 B & 8.42 B \\
    MPNet{*} & 8.42 B & 8.42 B & 8.42 B \\
    MLLaMA13B{*} & 1.75 T & 1.75 T & 1.75 T \\ 
    \hline
    \end{tabular}
    \caption{Total number of FLOPs required for a single forward pass. Here, "Anor", "Depr", and "SH" stand for anorexia, depression, and self-harm. Models marked with {*} are baselines.}
    \label{tab:flops_comparison}
\end{table}





To study the computational efficiency of our framework, we report the number of floating point operations (FLOPs) in a single forward pass \cite{kaplan2020scaling}. 
We observe that the maximum number of FLOPs in the proposed methodology is 25.5 million, while the minimum number of FLOPs in the considered baselines is 8.42 billion. This reduces the total number of floating point operations by 330 times in the worst-case (Appendix \ref{sec:flos-calculation}).

\begin{figure}
    \centering
\includegraphics[width=0.89\columnwidth]{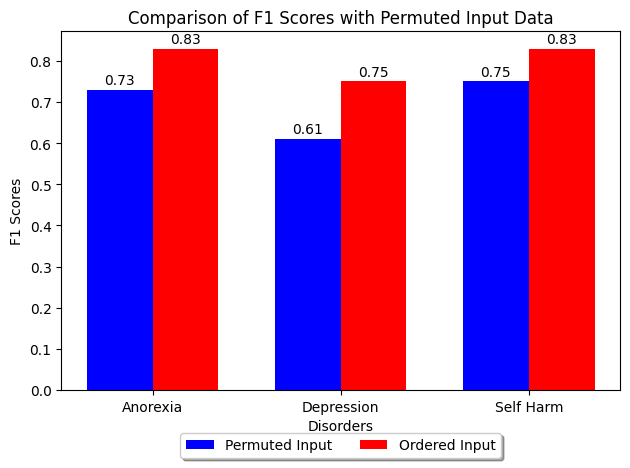}
    \caption{Results for temporal analysis: F1 scores comparison between the permuted input data and the ordered input data for three disorders of condition class}
    \label{fig:permuted}
\end{figure}
\subsection{Temporal Analysis}
\label{sec:temporal_analysis}
Temporal analysis is essential in our setup since anchor embedding was derived from condition class users. The cosine similarity values for condition classes are expected to be higher than the control class in the time-series representation (Figure \ref{fig:temporal_representation}). To understand if our model is picking up the temporal order (not just the magnitude-related information), we perform the training after permuting the input data five times in random order. This ensures that the post-wise temporal property of data is lost. From  Figure \ref{fig:permuted}, we observe a significant dip in performance as compared to our original setup and note that the performance is worse or comparable to \citet{aragon-etal-2023-disorbert}. These results empirically establish the importance of post-wise temporal properties of social media data.

\subsection{Ablation Study}
\label{sec:ablation_study}
In this section, we study the importance of anchor embedding by introducing three different ablation setups without an anchor:

\textbf{Direct Encoding:} This involves direct encoding of the posts by using MPNet \cite{agarwal2024analysing} and utilizes the complete 768 dimension vector without anchor embedding.

\textbf{Probabilities Values:} This involves calculating probability values using three Huggingface models, each fine-tuned for specific mental health conditions: a BERT \cite{devlin-etal-2019-bert} model for self-harm\footnote{\href{https://huggingface.co/dkuzmenko/self-harm-bert}{Self-Harm-HF}}, and RoBERTa \cite{liu2019roberta} models for anorexia\footnote{\href{https://huggingface.co/driwnet/Roberta_Anorexia_1}{Anorexia-HF}} and depression\footnote{\href{https://huggingface.co/ShreyaR/finetuned-roberta-depression}{Depression-HF}}. The probabilities are then arranged in chronological order for time series classification, without creating anchors.

\textbf{Plutchik-wheel-based Emotions:} This setup involves using eight different emotions (anger, anticipation, joy, trust, fear, surprise, sadness and disgust) based on Plutchik-wheel \cite{sawhney2021phase} for individual posts. Here, we used the probability scores of a post for each emotion and performed multivariate time series classification in a similar setup.
\begin{table}[htb!]
    \centering
    \small
    \begin{tabular}{c c c c}
    \hline
    \textbf{Model} & \textbf{F1} & \textbf{P} & \textbf{R} \\
    \hline
    \multicolumn{4}{c}{\textbf{Anorexia}}\\
    \hline
    Direct Encoding& 0.68 & 0.63 & 0.80\\
    Probabilities Values& 0.76 & 0.79& 0.73\\
    Plutchik-wheel-based Emotions& 0.30 & 0.20& 0.72\\
    LSTM (Anchor-based Approach) & \textbf{0.79} & 0.84 & 0.74\\
    \hline
    \multicolumn{4}{c}{\textbf{Depression}} \\
    \hline
    Direct Encoding& 0.66 & 0.64 & 0.70\\
    Probabilities Values& 0.47 & 0.46 & 0.48\\
    Plutchik-wheel-based emotions& 0.62 & 0.69 & 0.56\\
    LSTM (Anchor-based Approach) & \textbf{0.75} & 0.79 & 0.71\\
\hline
    \multicolumn{4}{c}{\textbf{Self-Harm}}  \\
    \hline
    Direct Encoding& 0.62 & 0.72 & 0.56\\
    Probabilities Values& 0.73 & 0.81 & 0.66\\
    Plutchik-wheel-based emotions& 0.47 & 0.37 & 0.67\\
    LSTM (Anchor-based Approach) & \textbf{0.83} & 0.93 & 0.75\\
\hline
    \end{tabular}						
    \caption{Ablation Study - Comparison of three alternative setups without anchor embedding}
    \label{tab:ablated_results}
    \vspace{-5pt}
\end{table}

We observe that the scores for various ablated model variants are consistently lower than those of our proposed anchor-based method, underscoring the importance of anchor embeddings.

\section{Full Context Analysis}
\label{sec:full-context}
\begin{figure}
    \centering
    \includegraphics[width=1\columnwidth]{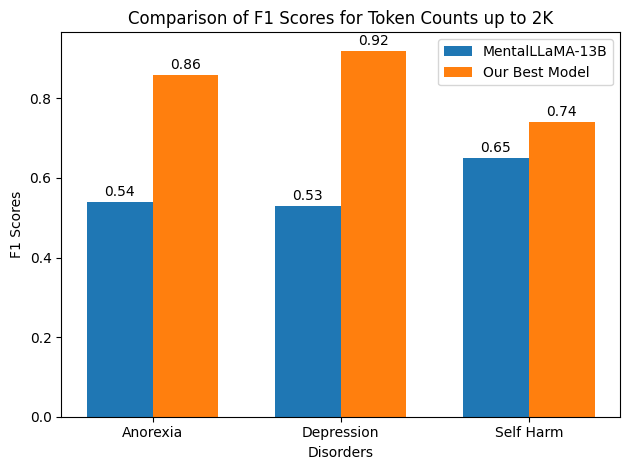}
    \caption{F1 scores of the condition class in three eRisk tasks by considering up to 2k context length.}
    \label{fig:full-context}
\end{figure}

In this section, we compare our approach with LLMs in the scenario when the data instances have context length within the capacity of the LLM. The main purpose of this experiment is to understand if LLMs can understand and capture the temporality directly from the natural language. We perform this experiment using MentalLLaMA \cite{yang2024mentallama} because of its superior performance in mental disorder identification over GPT and other LLMs.
For this experiment, we only consider the data instances with word count within the capacity of MentalLLaMA (i.e. 2048 tokens). We report the data statistics in the Table \ref{tab:data_statistics_2k}. It is important to note that users in the dataset whose histories fit within the context window may also systematically differ from the larger population, which could affect the generalizability of the results. Additionally, we have a very small number of instances of depression, so the results may not be conclusive for this case.
We report all the results in Figure \ref{fig:full-context}. We observe that our approach outperforms MentalLLaMA in all three tasks. This observation aligns with recent studies like \citet{liu2024lost}, which demonstrates the inability of LLMs to utilize long context inputs.

\section{Transfer-learning Setting}
\label{sec:cross-domain}

This section aims to study the similarity between linguistic cues shared across mental disorders by applying the proposed framework.

Various psychological studies (\citealp{smithuis2018self, lundh2011depressive, calvo2022comorbidity}) highlight the possible coexistence of multiple disorders in one person.
Overall, 25\%-55\% of patients with eating disorders showed self-harming behaviour \cite{raemen2020non}; at least 64\% of the patients with anorexia experience co-morbid major depressive disorder \cite{riquin2021severity} and  36.6\% of individuals with depression attempted suicide 
\cite{al2013assessment}. 

The similarity between anorexia and depression can be explained by their common behavioural symptoms like feelings of low self-esteem and greater self-criticism \cite{calvo2022comorbidity}. 
For example: \textit{``Feeling overwhelmed today. Skipped breakfast again; just couldn't face it. Trying to stay positive, but some days are harder than others.”}
This person shows depression and anorexia symptoms at the same time. The relationship between self-harm and depression can be understood by their shared symptoms like rumination, shame, guilt, regret, etc. \cite{lundh2011depressive}. For example, a post like \textit{``It’s hard to fight this guilt daily. I wish to forget this pain."}, shows that the person is feeling depressed and has a self-harming tendency. Notably, depression is commonly observed in patients diagnosed with self-harm and anorexia. However, the converse is not necessarily true. Anorexia and self-harm are both considered maladaptive coping mechanisms occurring when an individual can not figure out healthy ways to channel their thoughts, resulting in harming oneself by self-starvation \cite{smithuis2018self}.
\begin{table}[htb!]
    \centering
    \small
    \begin{tabular}{c c c c c c}
    \hline
    Model &Train+Val & Test &F1& P& R \\
    \hline
    \multicolumn{6}{c}{Anorexia (A)}\\
    \hline
    DisorBERT&  A &A& \textbf{0.83} & 0.82 & \textbf{0.85}\\
    LSTM&  A &A& 0.79 & 0.84& 0.74\\
    LSTM&  D &A& 0.75 & 0.68& 0.83\\
    LSTM& SH &A& 0.80&\textbf{0.85}&0.75\\
    \hline
    \multicolumn{6}{c}{Depression (D)} \\
    \hline
    DisorBERT&  D &D& 0.69 & 0.56 & \textbf{0.89}\\
    LSTM&   D&D& \textbf{0.75} & 0.79& 0.71\\
    LSTM&   A &D& 0.63& \textbf{0.86}& 0.50\\
    LSTM&   SH&D& 0.63& 0.73& 0.56\\
\hline
    \multicolumn{6}{c}{Self-Harm (SH)}  \\
    \hline
    DisorBERT&   SH &SH& 0.74 & 0.72 & 0.76\\
    LSTM&   SH&SH& \textbf{0.83}& \textbf{0.93}& 0.75\\
    LSTM&  A &SH& 0.78 & 0.85 & 0.72\\
    LSTM&  D &SH& 0.69&0.65 &\textbf{0.77}\\
\hline
    \end{tabular}
								
    \caption{Results for transfer-learning evaluations for all six combinations of disorders. Here, `A' is anorexia, `D' is depression and `SH' is self-harm.}
    \label{tab:cross_domain_results}
\end{table}

To understand the commonality of the disorder $\mathcal{D}_1$ with respect to disorder $\mathcal{D}_2$, we take the cosine similarity between the anchor embedding of the $\mathcal{D}_1$ disorder with the post embeddings of $\mathcal{D}_2$ disorder arranged in chronological order. After obtaining this series of cosine similarities (Section \ref{sec:Time Series Representation}), we use it as the train set and evaluate the test set of the disorder $\mathcal{D}_1$. This step aims to extract shared information about the mental disorder under consideration from the data of other disorders. For example, to study the linguistic similarity of depression with self-harm, we take the cosine similarity between the anchor embedding of depression and social media posts embedding from self-harm. Using the acquired series as the training set, we report the test set results from depression. We observe that the results are significantly better than the random baseline and competitive with the SOTA in certain cases. We present results on all three pairs (6 combinations) in Table \ref{tab:cross_domain_results}. Specifically, we observe that anorexia and self-harm show good F1 scores in the transfer-learning setting, whereas the other pairs involving depression show sub-optimal results as compared to SOTA. This may be due to the extra-linguistic features in the depressed class subjects, which may act as noise for the other disorders. While a substantial number of patients with anorexia or self-harm also experience depression, not every individual with depression exhibits symptoms of these mental disorders. Overall, this indicates that linguistic cues essential for classifying one disorder may be present in others, hinting at the potential of leveraging data for other domains.

\section{Conclusion and Future Work}

This work proposes a novel framework to incorporate temporal representation of textual data for the identification of Anorexia, Depression, and Self-harm from social media data.
Our methodology utilizes fundamental deep-learning architecture and surpasses LLM-based baselines by accounting for temporality and the full context of the input data. Our transfer-learning analysis highlights the overlapping linguistic cues among the disorders and hints at the possibility of leveraging data from different mental disorders.

Our work can be extended by exploring more complex mental disorders such as schizophrenia, personality disorders, bipolar disorder etc. The task of mental disorder classification can be enhanced by including other modalities like audio and visual signals that give insights into behavioural patterns.
 The proposed framework can be also extended beyond the mental health domain in scenarios which require a temporal understanding of natural language data. 

\section*{Limitations}
Our study's limitations arise from the challenges inherent in analyzing mental disorders through social media data. Notably, as discussed in our error analysis, the presence of out-of-context posts and individuals' reluctance to openly express their mental health challenges due to stigma can lead to false negatives. Additionally, the incomplete context within social media posts, where some content is removed or censored for community well-being, can hinder the framework's accuracy in identifying mental disorders, emphasizing the complexities of using online data. Other limitations of this work may arise from the use of traditional deep learning architectures. For instance, the feed-forward and CNN networks were sensitive to the choice of hyperparameters, especially for anorexia. To handle this, an extensive grid search was performed.  

\section*{Ethics Statement}
When conducting an analysis of social media content pertaining to mental disorders, it is essential to address valid concerns regarding individual privacy and ethical considerations. These concerns stem from handling sensitive and personal information, including discussions about emotions and health-related issues. 
All the examples of social media posts mentioned in this paper are \textit{paraphrased and anonymized representations of the actual data} as we do not have permission to publish any portion of the dataset (e.g. example post) other than summary statistics. It is worth noting that we have exclusively utilized publicly available datasets, specifically Reddit datasets (Section 3.1) and eRisk collections (Section 4.1). We have diligently adhered to the terms of use and \href{https://tec.citius.usc.es/ir/code/eRisk2019.html}{user agreements} associated with these collections. Furthermore, the datasets we have employed are anonymized, and our research does not involve any direct interaction with social media users. Given these conditions and practices, our study does not necessitate review and approval by an Ethics Committee Board.
\bibliography{acl2023}
\bibliographystyle{acl_natbib}

\appendix

\section{Basic Definitions}
\label{appendix-definitions}
We briefly define all the considered disorders discussed in Section \ref{sec:introduction}. 
\paragraph{Anorexia nervosa:}A serious eating disorder characterized by extreme self-starvation and weight loss, leading to a low body weight. 
    
\paragraph{Depression:}A mental disorder characterized by persistent sadness and a lack of interest in previously rewarding or enjoyable activities.
    
\paragraph{Self-harm:}A deliberate act of inflicting harm upon oneself which can include cutting, scratching, or hitting oneself.


    
\begin{table}[htbp]
    \centering
    \small
    \begin{tabular}{cccc}
    \hline
    &  r/ED &  r/depr & r/suicide \\
    \hline
    total \# posts & 9535 & 58089 & 41354 \\
    avg \# words &  129.59 & 190.69 & 171.25 \\
    \hline     
    \end{tabular}
    \caption{Statistics of the RMHD dataset. "r/ED" and "r/depr" stand for eating disorder and depression-specific subreddits respectively
    }
    \label{tab:RMHD}
\end{table}

\section{Baseline Approaches}
\label{sec:baseline_appendix}
As discussed in Section \ref{sec:baseline_approaches}, we consider the following baselines from \citet{aragon-etal-2023-disorbert}:

\begin{table*}[htbp]
    \centering
    \begin{tabular}{cccccccccc}
    \hline
Method &\multicolumn{3}{c}{Anorexia} & \multicolumn{3}{c}{Depression} & \multicolumn{3}{c}{Self-Harm} \\
    \hline
    &  F1 & P & R & F1 & P & R & F1 & P & R \\
    \hline
    Decision Tree & 0.66 & 0.66 & 0.66 & 0.54 & 0.74 & 0.42 & 0.69 & 0.71 & 0.67 \\
    XGBoost       &  0.74 & 0.83 & 0.67 & 0.55 & 0.77 & 0.42 & 0.74 & 0.86 & 0.64 \\
    Adaboost      & 0.74 & 0.81 & 0.68 & 0.50 & 0.88 & 0.35 & 0.78 & 0.87 & 0.71 \\
    Random Forest &  0.75 & 0.86 & 0.67 & 0.57 & 0.85 & 0.42 & 0.78 & 0.84 & 0.72 \\
    LightGBM      & 0.81 & 0.86 & 0.77 & 0.53 & 0.88 & 0.38 & 0.83 & 0.90 & 0.77 \\
   
    \hline
    \end{tabular}

\caption{Results obtained by performing experiments using various machine learning models. F1, precision (P), and recall (R) values over the condition class are reported for the three tasks: Anorexia, Depression and Self-Harm. }
\label{tab:ERISK-RESULTS-ml}
\end{table*}



\textbf{BERT} \cite{devlin-etal-2019-bert}: This approach involves implementing a BERT-based model fine-tuned for adapting to a specific training set.

\textbf{MentalBERT} \cite{ji-etal-2022-mentalbert}: This pre-trained language model is tailored for the mental healthcare domain and constructed using an extensive dataset of sentences sourced from Reddit. Similar to the fine-tuning process employed for BERT, we adapted this model to each training set.

\textbf{DisorBERT} \cite{aragon-etal-2023-disorbert}: This double-domain adapted language model is tailored to mental healthcare, similar to MentalBERT. Initially, BERT was fine-tuned to capture the language structure commonly found on large social media platforms like Reddit. Subsequently, further adaptation was made to specialize the model in understanding the specific language used by individuals with mental disorders.


Additionally we also use MPNet \cite{reimers2019sentence}, GPT-3.5-turbo \cite{brown2020language} and MentalLLaMa-chat-13B \cite{yang2024mentallama} as our baselines. They are described below:

\textbf{MPNet (Zero-shot):} We use the \textit{all-mpnet-base-v2} model in a zero-shot manner on the test set of the e-Risk dataset for all the three tasks. 

\textbf{MPNet (Fine-Tuned):} We fine-tune the \textit{all-mpnet-base-v2} model using the train and val set of the e-Risk datasets for all the three tasks and evaluate the performance on their respective test sets.

\textbf{GPT-3.5-turbo:} We prompt the GPT-3.5 Turbo in a zero-shot manner on each chunk of the test set individually, followed by a majority voting mechanism for binary classification across all three tasks.

\textbf{MentalLLaMa-chat-13B:} For MentalLLaMa-chat-13B, we utilized a zero-shot prompting approach, presenting each chunk of the test set individually, followed by a majority voting mechanism for binary classification across all three tasks.

For GPT-3.5-turbo and MentalLLaMa-chat-13B models, we used the following prompt:
\textit{"Consider this post: \{chunk text\}. Does the poster suffer from Anorexia/Depression/Self-Harm? You are instructed to answer in YES or NO."}

\section{Model Configurations and Experiments}
\label{sec:appendix-model-config-baseline}

As discussed in Section \ref{subsec:Time-series-Classification}, we describe the model configuration and experiments here.

The Feedforward network and ML models were trained on the top 30 \footnote{list of extracted features can be found \href{https://inquisitive-pancake-bcb.notion.site/Extracted-Features-using-tsfresh-f063eda880c94677932f7f494902b099?pvs=4}{here}}  selected features extracted from time series data. For feature selection, we employed the Gini impurity criterion using a Random Forest classifier, as described by \citet{yuan2021gini}. We selected the top 30 features based on their Gini importance scores to serve as independent variables for learning the decision space of mental disorders. Time-Series Transformer, 1D-CNN, and LSTM-based classifiers were trained on the temporal representation of data. We used the Adam optimizer, cross-entropy, as a loss function for all our experiments. In our experimental setup, we conducted training using a single NVIDIA A100-SXM4-80GB GPU. During training, we employed callbacks for model checkpointing and early stopping to optimize and prevent overfitting. We use a grid search on the validation set to search for optimal hyperparameters. 
Table \ref{tab:hyperparameters} provides details of the hyperparameters used for each model in the context of Anorexia, Depression, and Self-Harm classification tasks. These hyperparameters include the learning rate (lr), batch size (\# BS), and the number of epochs (\# E) for each deep learning model.

\subsection{Additional Experiments}
\label{sec:appendix-additional-exp}

Apart from the baseline mentioned in Section \ref{sec:baseline_appendix}, we also perform experiments on MPNet-base \cite{song2020mpnet}, RoBERTa-base \cite{liu2019roberta} and DeBERTa-base \cite{he2020deberta} models in zero-shot and fine-tune settings. For finetuning, we use the train and validation set of the e-Risk datasets for all three tasks. The results are shown in the Table \ref{tab:additional_exp}. We observe an improvement after finetuning the models but the numbers are still below the current SOTA.  

We also perform experiments on various machine learning models. Table \ref{tab:ERISK-RESULTS-ml} summarizes the results of these experiments for the tasks of Anorexia, Depression, and Self-Harm classification.


\label{sec:appendix-hyperparameters}

\begin{table*}[htbp]
    \centering
    \begin{tabular}{cccccccccc}
    \hline
Method &\multicolumn{3}{c}{Anorexia} & \multicolumn{3}{c}{Depression} & \multicolumn{3}{c}{Self-Harm} \\
    \hline
    &  lr & \# BS & \# E & lr & \# BS & \# E & lr &  \# BS & \# E \\
    \hline
    Feedforward Network &  1e-3 & 16 & 200  & 1e-4 & 32 & 200 & 1e-3 & 2 & 200 \\
    Time-Series Transformer &  3e-4 & 16 & 20 & 3e-4 & 32 & 10 & 1e-3 & 16 & 20 \\
    1D-CNN &  1e-3 & 16 & 50 & 1e-3 & 8 & 100 & 1e-3 & 16 & 100 \\
    LSTM &  1e-2 & 8 & 50 & 1e-2 & 16 & 50 & 1e-2 & 8 & 50 \\
    \hline
    \end{tabular}

\caption{Experimental hyperparameter values for each model across all the three tasks. Here, lr represents learning rate, \# BS represents batch size, and \# E represents number of epoch.}
\label{tab:hyperparameters}
\end{table*}

\section{Langauge Model and Context Length}
\label{sec:appendix_context_lengths}
As discussed regarding the maximum token length handled by language models in section \ref{sec:Time Series Representation}, models like BERT \cite{devlin-etal-2019-bert}, MPNet \cite{song2020mpnet}, T5 \cite{raffel2020exploring} have a maximum context length of 512 with BART \cite{lewis2019bart} and Vicuna \cite{chiang2023vicuna} having 1024 and 2 K context length respectively. 
LLMs like GPT-3.5 \cite{brown2020language}, LlaMA-2 \cite{touvron2023llama}, and Zephyr \cite{tunstall2023zephyr} have a 4 K context length. Contemporary LLMs like Mistral \cite{karamcheti2021mistral}, DeepSeek LLM \cite{bi2024deepseek}, and Qwen 1.5 \cite{bai2023qwen} have 8K, 32K, and 32K context lengths, respectively.
We have approximately 14.4 K words for depression, 16.2 K words for anorexia, and 6 K words for self-harm per subject on average. For processing a long social media post history, the resulting token length will exceed the full context length of the contemporary language model. 
 

\section{FLOPs Calculation}
\label{sec:flos-calculation}
As discussed in Section \ref{sec:efficiency-analysis}, we followed the methodology proposed by \citet{kaplan2020scaling} for approximating the number of floating point operations (Flos) in transformer ($C_{\text{forward}} \approx 2N + 2n_{\text{layer}} \cdot n_{\text{context}}  \cdot d_{\text{model}}$). The models trained using our proposed framework require a comparatively lower number of Flos. Notably, the Feedforward and LSTM models only utilize Flos in order of thousands for a forward pass.

\begin{table*}[htbp]
    \centering
    \begin{tabular}{lccccccccc}
    \hline
Model &\multicolumn{3}{c}{Anorexia} & \multicolumn{3}{c}{Depression} & \multicolumn{3}{c}{Self-Harm} \\
    \hline
    &  F1 & P & R & F1 & P & R & F1 & P & R \\
    \hline
    MPNet (ZS)  & 0.16 & 0.09 & 1.00 & 0.62 & 0.45 & 1.00 & 0.40 & 0.25 & 1.00 \\
    MPNet (FT on eRisk) &  0.53 & 0.37 & 0.90 & 0.71 & 0.58 & 0.93 & 0.67 & 0.82 & 0.57 \\
    MPNet (FT on eRisk+RMHD) &  0.20 & 0.11 & 0.99 & 0.61 & 0.44 & 0.97 & 0.42 & 0.27 & 0.98 \\
    DeBERTa (ZS)      & 0.16 & 0.09 & 1.00 & 0.62 & 0.45 & 1.00 & 0.40 & 0.25 & 1.00 \\
    DeBERTa (FT on eRisk) &  0.73 & 0.62 & 0.89 & 0.59 & 0.61 & 0.57 & 0.36 & 0.86 & 0.23 \\
    DeBERTa (FT on eRisk+RMHD) & 0.23  & 0.13 & 0.99 & 0.61 & 0.44 & 0.97  & 0.42  & 0.27 & 0.98 \\
    RoBERTa (ZS)      & 0.16 & 0.09 & 1.00 & 0.62 & 0.45 & 1.00 & 0.40 & 0.25 & 1.00 \\
    RoBERTa (FT on eRisk)      & 0.68 & 0.55 & 0.89 & 0.63 & 0.59 & 0.68 & 0.36 & 0.96 & 0.22 \\
    RoBERTa (FT on eRisk+RMHD)      & 0.23 & 0.13 & 0.99 & 0.61 & 0.44 & 0.97 & 0.42 & 0.27 & 0.97 \\
   
    \hline
    \end{tabular}

\caption{Results of additional experiments on MPNet-base, RoBERTa-base and DeBERTa-base models. F1, precision (P), and recall (R) values are reported over the condition class in three e-Risk tasks: Anorexia, Depression and Self-Harm. ZS and FT refer to Zero-Shot and Fine-Tuned experiments respectively.}
\label{tab:additional_exp}
\end{table*}

\section{Limited Performance of GPT-3.5-turbo and MentalLLaMA-chat-13B}
\label{sec:poor-performance-gpt3.5-mentalLLaMa}
As discussed in the Section \ref{sec:Results-and-Observation}, we describe the responses generated by GPT3.5 and Menatallama-13B here.
The responses from the following prompts show that the individual chunk does not contain enough information to determine if the poster suffers from a mental disorder. Therefore, the response from both GPT-3.5-turbo and MentalLLama-13B-chat in most of the instances is "NO". This results in a high false negative and low F1 score for the condition class, as seen in Table \ref{tab:ERISK-RESULTS}. We discuss an example explaining this situation here.

\paragraph{Prompt:}
Consider this post: \textit{"Years will pass, and I can't recall anything that happened. The Wheel of Time series is a great option for a long-term, mid-fantasy escape. Holy sh*t, dude"} Question: Does the poster suffer from Depression?

\paragraph{Response from GPT-3.5-turbo:} Based solely on the provided post, \textbf{it's difficult to conclude whether the poster suffers from depression} The post seems to reflect a sense of detachment or disconnection from the passage of time and memory, but it's unclear if this is indicative of depression or simply a passing observation.

\paragraph{Response from MentalLLama-13B-chat:} The post \textit{\textbf{does not exhibit strong emotional indicators of very severe depression}}. It mentions having no memory of what transpired, which could potentially be a symptom of depression, but it does not provide enough information to suggest a deep sense of emotional pain, exhaustion, or numbness.

\paragraph{Explanation} 
In the example, we can observe that the text is a concatenation of multiple post. The phrase \textit{"Years will pass, and I can't recall anything that happened."} belongs to the context where the subject was talking about depression, and the sentence \textit{"The Wheel of Time series is a great option...."} belongs to the post which talked about some irrelevant information. We can make the following observation from the prediction of GPT-3.5 and MentalLLama-13B: 1) the semantic significance of the first phrase is diminished due to the addition of the second phrase. 2) The first phrase itself can have ambiguity and is not enough to draw a conclusion about the depression. The subject may be talking about some other topics. Overall, these reasons contribute collectively to the misclassification of the disorder.

\section{Full Context Analysis}
\label{appendix:full-context_appendix}
\begin{table}[htbp]
    \centering
    \begin{tabular}{ccc}
    \hline
         & Condition &  Control \\
    \hline
    &\textbf{Anorexia}&\\
    \hline
    \#subjects &  16 &  101\\
    \hline
    &\textbf{Depression}&\\
    \hline
     \#subjects&  7&  8\\
 
    \hline
    &\textbf{Self-Harm}&\\
    \hline
    \#subjects&  51 &  126\\
    \hline
    \end{tabular}
								
    \caption{Statistics of the e-Risk datasets for anorexia, depression and self-harm considering the total of 2k context length. The \textit{control} class refers to the class of individuals not diagnosed with a disorder and the \textit{condition} class refers to the class of individuals diagnosed with a disorder.}
    \label{tab:data_statistics_2k}
    
\end{table}
The dataset utilized for full context analysis, as described in Section \ref{sec:full-context}, consists of data instances with word counts within the capacity of 2048 tokens, suitable for the MentalLLaMA model. The Anorexia condition group includes 16 subjects, while the control group comprises 101 subjects. The Depression condition group has a relatively smaller sample size, with 7 subjects in the condition group and 8 in the control group. The Self-Harm condition group has a larger sample, with 51 subjects in the condition group and 126 in the control group.

\end{document}